\documentclass{mva_style}
\usepackage{graphicx}
\usepackage{tikz}
\usepackage{amssymb,amstext,amsmath}
\usepackage{algorithm}
\usepackage{algpseudocode}
\usepackage{sidecap}
\usepackage{url}
\usepackage[sort]{cite}

\usepackage{hyperref}
\hypersetup{
    colorlinks,%
    citecolor=blue,%
    filecolor=black,%
    linkcolor=red,%
    urlcolor=black
}

\makeatletter
\def\hlinewd#1{%
\noalign{\ifnum0=`}\fi\hrule \@height #1 %
\futurelet\reserved@a\@xhline}
\usepackage{tikz}

\finalcopy %Uncomment this line for the Camera-Ready Manuscript

\begin{document}
\title{Handwritten and Printed Text Separation in Real Document}
\author{
  Abdel Bela\"id, K.C. Santosh\\
  LORIA - Universit\'e de Lorraine\\
  54506 Vandoeuvre-l\`es- Nancy, France\\
  {\tt \{abdel.belaid, santosh.kc\}@loria.fr}\\
  \and
  Vincent Poulain d'Andecy\\
  ITESOFT Parc d’Andron, Le S\'equoia,\\
  30470, Aimargues, France\\
  {\tt vincent.poulaindandecy@itesoft.com}\\
}

\maketitle

\section*{\centering Abstract}
\textit{The aim of the paper is to separate handwritten and printed text from a real document embedded with noise, graphics including annotations. Relying on run-length smoothing algorithm (RLSA), the extracted pseudo-lines and pseudo-words are used as basic blocks for classification. To handle this, a multi-class support vector machine (SVM) with Gaussian kernel performs a first labelling of each pseudo-word including the study of local neighbourhood. It then propagates the context between neighbours so that we can correct possible labelling errors. Considering running time complexity issue, we propose linear complexity methods where we use $k$-NN with constraint. When using a kd-tree, it is almost linearly proportional to the number of pseudo-words. The performance of our system is close to 90\%, even when very small learning dataset are used, where samples are basically composed of complex administrative documents.
}

\section{Introduction}
Under the purview of document analysis and processing, we are in this paper, motivated to separate handwritten and machine-printed text ($\cal H\& \cal P$) so that further processing is feasible such as document information exploitation and retrieval. In other words, such a separation is an important step in the process because it allows retro-conversion to avoid heavy treatments and errors when transcribing the content. 

Considering a continuous flow of administrative documents into our system, we face a varieties of document types, content, quality and structure. Fundamentally speaking, documents can be skewed, noisy and sometimes overlapped with graphics i.e., lines and unconstrained annotations. In this context, most of the image samples are required to be properly treated. Without integrating such tools, our system, in this framework, aims to extract the annotations whatever the language: French, German and English used in the document, the content: typed or handwritten, and document structure: structured (e.g. tables), semi-structured (e.g. forms) and structure-free. Although the segmentation topic has been studied since several
years~\cite{Kang}, different methods have been proposed to solve particular aspects of the separation~\cite{Chanda10,setlur11}. Heterogeneous document separation still remains an open problem. Another strong industrial constraint is to reduce running time so that the system can maintain speed. In addition, parameter-free methods are always better since they can generally be applied. In this paper, we are motivated by the work of Kandan et al.~\cite{Kandan} where separation has been made into two classes by using descriptors that are insensitive to translation, rotation and scaling. Classifications using SVM and $k$-NN are first investigated, and a re-classification step is then performed using a Delaunay triangulation. Zheng \textit{et} al. proposed two segmentation approaches and evaluated over noisy documents~\cite{Zheng}. The first one is used to determine the most appropriate segmentation where a comparison is made between the segmentation into words, lines and connected components. The latter one deals with word classification by selecting 31 descriptors over a hundred. They also introduce information about class in order to take the noise into account. Fisher classifier is used to label the segmented blocks and Markov field then allows fine classification, considering the contextual information of each word. 

The rest of this paper is organised as follows. We start with detailing our proposed approach in Section~\ref{sec: proposedM}. It mainly includes pre-processing, pseudo-word segmentation, word model training, word classification and pseudo-word grouping. Full experimental results (and of course, analysis) are reported in Section~\ref{sec: exp}. The paper is concluded in Section~\ref{sec: concl} including a few perspectives.

\section{The proposed approach}\label{sec: proposedM}

\begin{figure}[tbp]
\centering
\includegraphics[scale=1]{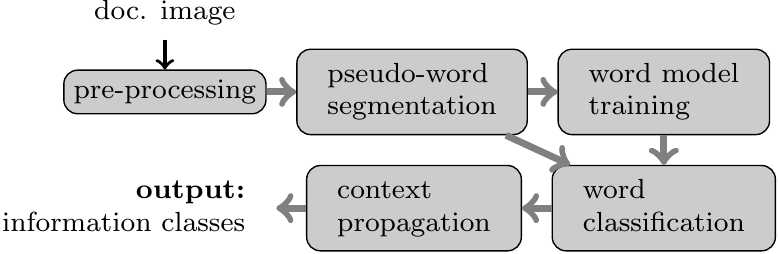}
\caption{Work-flow showing several consecutive stages, starting from pre-processing to output i.e., $\cal H\& \cal P$ text separation.}\label{block}
\end{figure}

As illustrated in Fig.~\ref{block}, our proposed approach consists of several consecutive steps. It includes pre-processing, pseudo-word segmentation, word model training, word classification and context propagation. In what follows, we explain them, one after another.
\bigskip

\noindent \textit{{\textbf{Preprocessing.}}} %\label{sec:preprocess}
The low quality documents require a significant preprocessing. Our pre-processing is composed of the following steps:
\begin{enumerate}\itemsep-0.5em
\item edge removal by using a rule system based on shape and position of the connected components (CC); 
\item noise filtering by using a modified kfill~\cite{Chinnasarn}; 
\item slope detection by using the RAST method~\cite{vanBeusekom}; and 
\item filtering by using the modified k-fill on the de-skewed document. 
\end{enumerate}

\begin{figure}[tbp]
\centering
\small
\renewcommand{\tabcolsep}{0em}
\begin{tikzpicture}
\node[-, draw, gray!50] (b) at (0,0) 
{
{{\includegraphics[width=.45\linewidth]{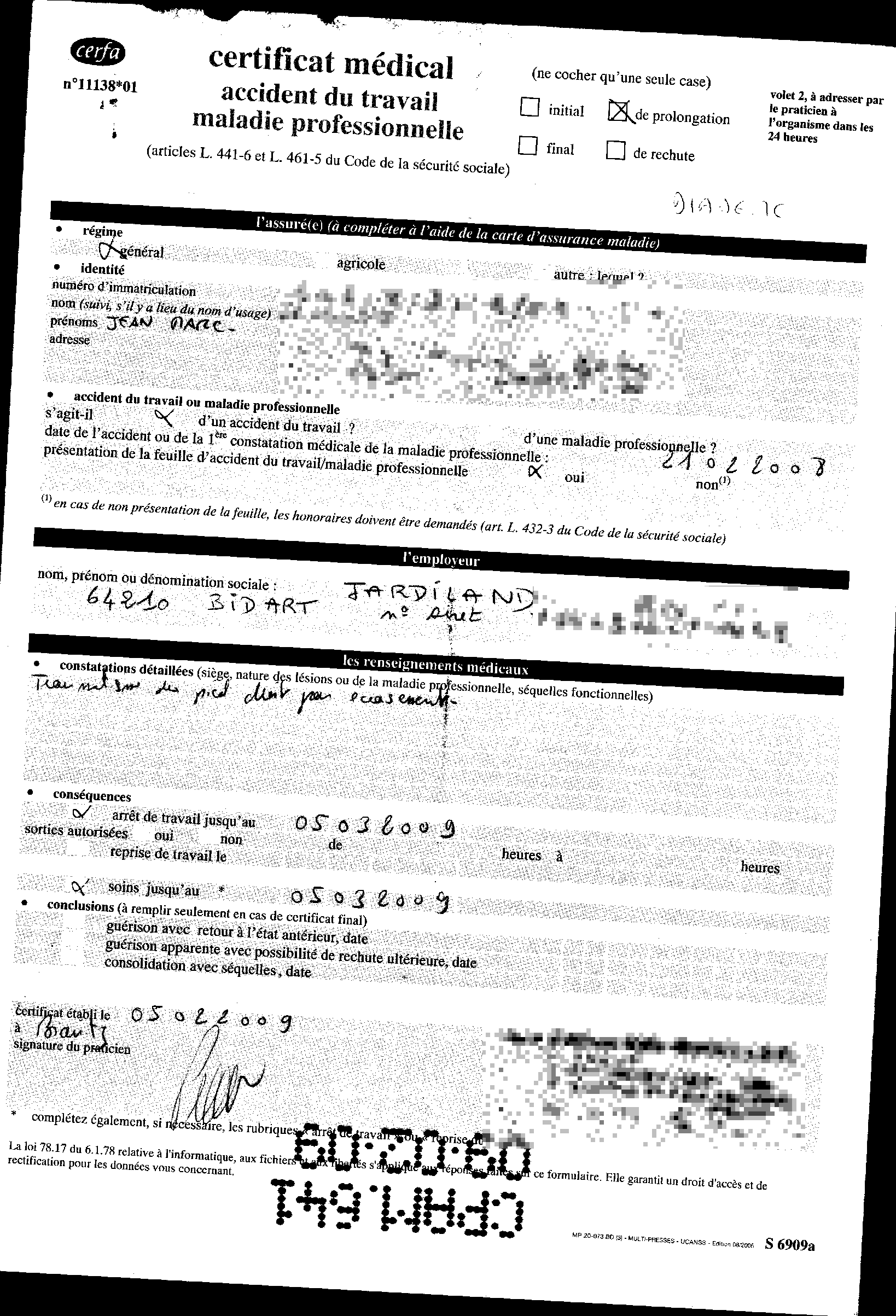}}} \hspace*{0.2em}
{{\includegraphics[width=.45\linewidth]{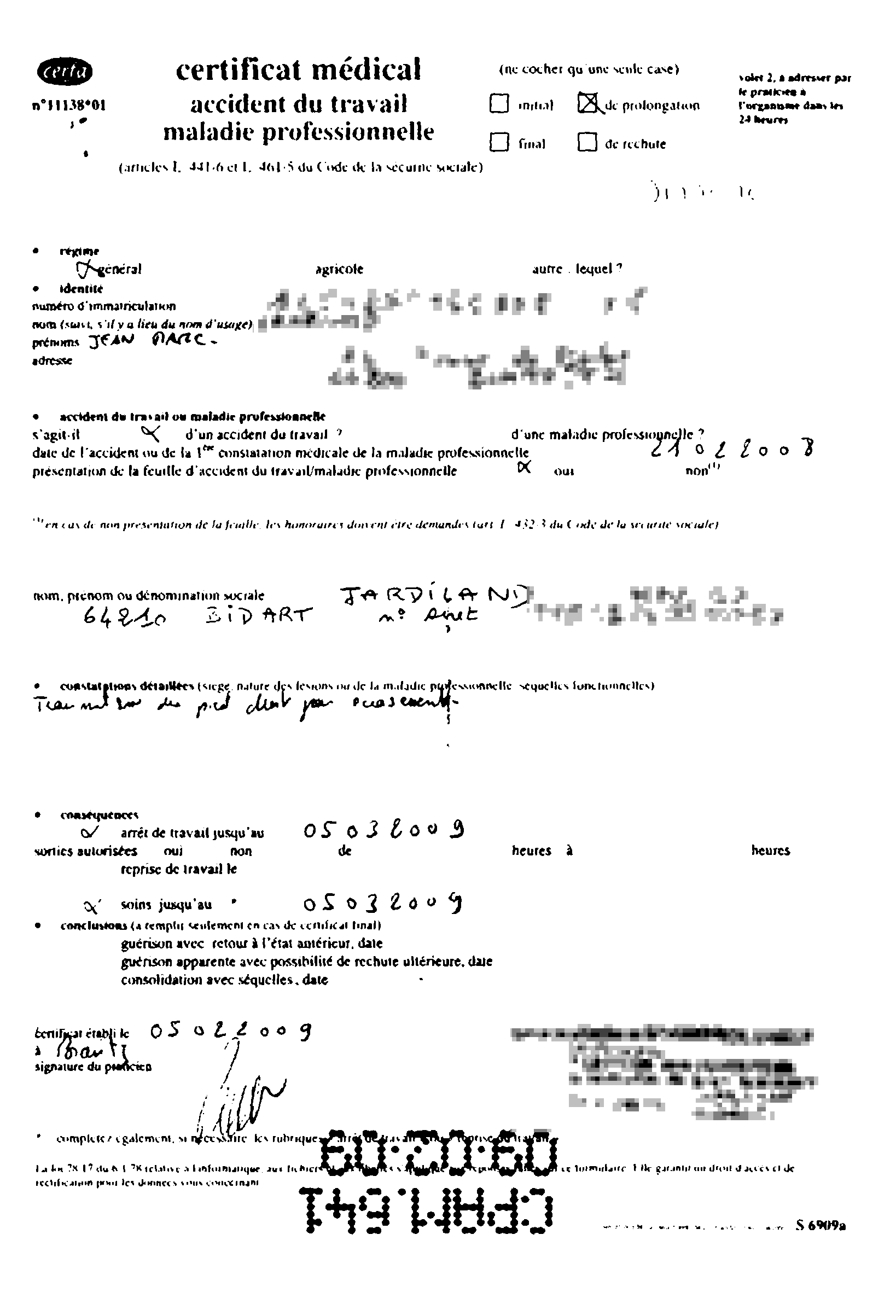}}}
};
\node at (-3.8,-2.5){(a)};
\node at (0.2,-2.5){(b)};
\end{tikzpicture}
%\begin{tabular}{c} {\fbox{\includegraphics[width=.45\linewidth]{05_flou}}} \\ (a) \end{tabular}
%\begin{tabular}{c} {\fbox{\includegraphics[width=.45\linewidth]{05_fltr_flou}}} \\(b) \end{tabular}\\[1em]
\caption{An example showing pre-processing: (a) input sample and (b) its corresponding output.}
\label{fig:preproc}
\end{figure}

\bigskip
\noindent \textit{\textbf{Pseudo-word segmentation.}} %\label{sec:wordSeg}
In this section, we create regular and stable areas that will be used to label the $\cal H\& \cal P$ zones in the document image. To handle this, we use a double RLSA as presented in Algorithm\ref{algo:doublesmear} i.e., it aims to provide fine word segmentation. 

\begin{figure}[tbp]
\centering
\small
\begin{tikzpicture}
\node[-, draw, gray!50] (b) at (0,0) 
{\includegraphics[width=0.95\columnwidth]{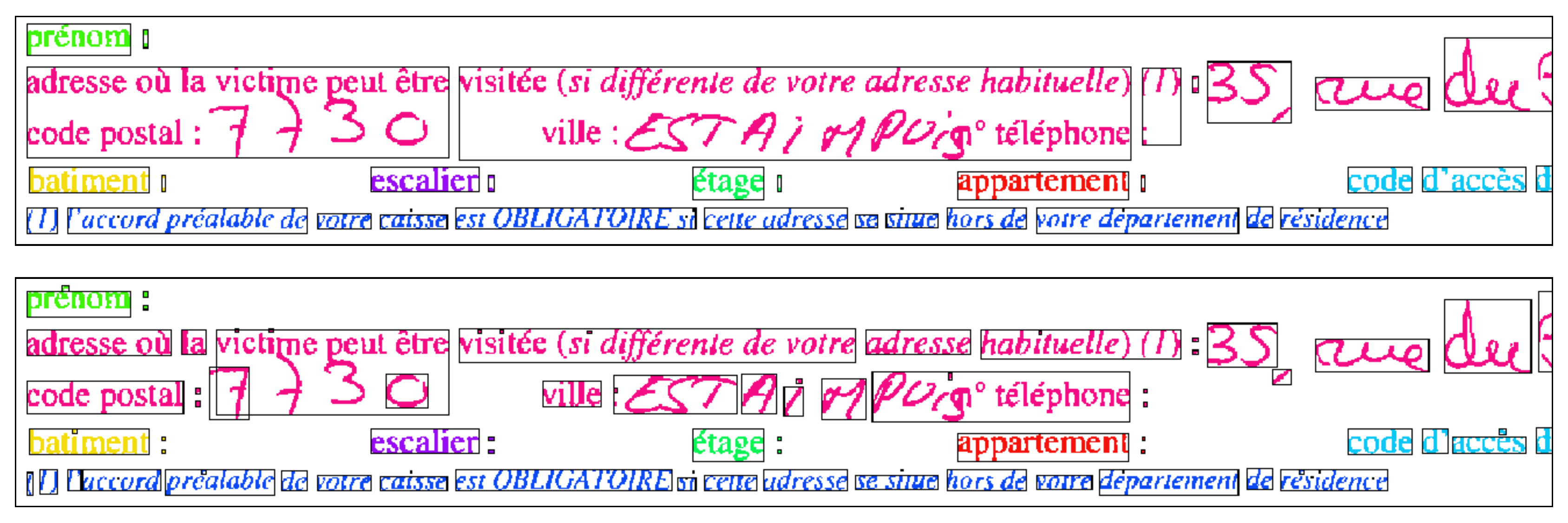}};
\node at (-4,0.5){(a)};
\node at (-4,-0.8){(b)};
\end{tikzpicture}
\caption{Segmentation comparison: (a) classical RLSA and (b) double RLSA. Extracted pseudo-words are framed and lines are identified by the color of the pseudo-words.}
    \label{fig:comparaisonseg}
\end{figure}

In each one of the extracted lines, smearing is performed first and the distances between the bounding boxes of the adjacent CC are then calculated. This allows to construct a histogram that generally provides an overall shape appearance. It contains two dominant peaks: 
\begin{enumerate}\itemsep-0.5em
\item the first corresponds to the most frequent gap between CC that can be considered as the distance between characters of the same word; and 
\item the second peak corresponds to the most frequent gaps between words belonging to the same row. 
\end{enumerate}
Note that the first peak can be considered as the distance between the letters in every word and in a similar fashion, the second peak determines the threshold to be used in pseudo-word segmentation. We can therefore apply a second smearing that allows a finer segmentation because handwritten and printed words do not respect similar (usual) distances between the letters and words, and thus we are able to adapt the row content segmentation. Fig.~\ref{fig:comparaisonseg} illustrates the comparison between the original and the double RLSA. In this illustration, it is important to notice that words are well segmented in case when double RLSA is used in contrast to text block (that sometimes contains several words within it) from classical RLSA.   
%%%%%%%%%%%%%%%%%%%%%%%%%%%%%%%%%%%%%%%%%%%%%%%%

\begin{algorithm}[tbp]
\footnotesize
\caption{Segmentation by double smearing}
\label{algo:doublesmear}
\begin{algorithmic}[1]
%\REQUIRE $\forall c \in \mathcal{C}\:, \ \quad \mathrm{old\_label}(c) \in \mathcal(L)$
\State $\mathit{lines} \gets smearing{(\mathit{image})}$
\ForAll{ line $L$ in $\mathit{lines}$}
	\State $\mathit{list\_edistances} \gets \emptyset$
	\ForAll { CC $c$ in line $L$}
		\State $d_{\mathit{min}} \gets distmin(\mathit{listeccx}, c)$
		\State $\mathit{list\_edistances} \gets add(\mathit{list\_edistances}, d_{\mathit{min}})$
	\EndFor
	%\State \Comment{calcul du nombre d'occurences de la valeur 0, 1, 2, etc.}
	\State $\mathit{compt} \gets bincount(\mathit{list\_edistance})$
	%\State \Comment{regroupement des valeurs 2 par 2, à partir de la troisième}
	\State $\mathit{histo} \gets \mathit{compt}[2::2] + \mathit{compt}[3::2] $
	\State $i \gets argmax(\mathit{histo})$
	\Repeat
 		\State $\mathit{previous} \gets \mathit{histo}[i]$
 		\State $i \gets i + 1$
	\Until { $\mathit{histo}[i] > \mathit{previous}$}
	\State $d_{hs} \gets i + 2 $
\EndFor
\end{algorithmic}
\end{algorithm}

\bigskip
\noindent \textit{\textbf{Word model training.}} %\label{sec:wordMod}
As said before, we need reliable models to separate $\cal H\& \cal P$ information. In order to have these models, we perform two classes of learning from samples by taking words representatives. We then select several specific descriptors belonging to four different categories: 
\begin{enumerate}\itemsep-0.5em
\item morphological (local properties of pseudo-words such as height, width and pixel number);
\item CC descriptors (11 descriptors as proposed in~\cite{Zheng}); 
\item pixel repartition (global descriptors like invariant HU moments, variance of the projection profiles~\cite{Kandan,daSilva}; and 
\item other local properties such as run length, crossing count and bi-level co-occurrences, as described in~\cite{Zheng}. %All descriptors form a vector of 137 values. 
\end{enumerate}

\bigskip 
\noindent \textit{\textbf{Classification.}} %\label{sec: class}
To handle pseudo-word classification, we employ a SVM. Although it is initially suggested to separate only the $\cal H\& \cal P$ information, we use a multi-class SVM so that an additional class i.e., noise can be taken into account. To handle this, two approaches are basically used: 1) the combination of bi-class SVM and 2) the learning of a unique multi-class SVM (MSVM). MSVM is based on a principle similar to one-vs-all~\cite{Vapnik} where each class has its own decision function and the class corresponding to the function giving the highest value wins. The difference is that, for a MSVM with $Q$ classes, the $Q$ functions are learnt at the same time with exactly similar constraints. A single optimization problem is solved by using the maximization of the sum of the margins for each class. There are four different methods that differ in terms of application penalty. We use the tool presented by Weston and Watkins~\cite{Weston} where it cumulates the penalty compared to the margins of each class. The implementation is carried out on the \textit{Weka} platform and the SMO classifier with the extension of the problem into three classes by the method one-vs-one as described in Mayoraz et al.~\cite{Mayoraz}.

\bigskip
\noindent \textit{\textbf{Pseudo-word grouping.}} This re-grouping method uses spatial proximity to re-group elementary units. For each component, $k$ nearest neighbours are found and the label of the component is compared with the ones in their neighbours. If more than 50\% of the neighbours share the same label, this label is assigned to the central component.

Generally speaking, since text is written horizontally, horizontal proximity between components is preferred to be vertical ones. Then, we define the distance as
\begin{equation}
d(e_1,e_2)=\sqrt{(x_1-x_2)^{2}w_{x}^{2} + (y_1-y_2)^{2}w_{y}^{2}}
\end{equation}
~\\
where $x_i, y_i$ are the coordinates of the center of gravity of CC $n_i$, and $w_{x;y}$ are weights corresponding to each axis. In a similar manner, another distance is computed i.e., the distance is taken from the border of the bounding boxes. Based on the framework, in what follows, we explain three different algorithms i.e., A1:A3.

\smallskip
\noindent {A1. \textit{Grouping by $k$-NN.}}\\
It employs a classical $k$-NN algorithm where parameters $k$ and a threshold i.e., {\textit{max}\_dist}. The $k$ nearest neighbours are taken into account if they are closer than the pre-defined \textit{max}\_dist. The distance parameter basically prevents far away neighbours to interfere with the component. In our case,\textit{max}\_dist. has been fixed to 1) 300 pixels for distance 1, and 2) 100 pixels for distance 2 with images at 300 dots per inch (dpi). Note that the distance 2 is lower than distance 1, and depends of the relative positioning between the bounding boxes and their sizes. These thresholds however, are image resolution dependent.

\smallskip
\noindent {A2. \textit{Grouping by the NN with constraints.}} \\
The algorithm can be improved by avoiding big components that are basically be corrupted by small ones (as noise). Before flipping the label of the component, we perform a test to check whether the accumulated pixels of a neighbour contributing the change of label is significant in comparison to the number of pixels of the tested component. For this, in our test, the sum should be at least 50\% of the main component. Note that the opposite does not exist. Big components are regrouped with small ones to help gathering main text with small components as commas, apostrophes or accents. Moreover, big components contain more information so they are generally more reliable, and thus the classification is more accurate. An overall idea is presented in Algorithm~\ref{algo:kNN1}.

\begin{algorithm}[tbp]
\footnotesize
\caption{$k$-NN grouping with constraints}
\label{algo:kNN1}
\begin{algorithmic}[1]
\\
{\bf Require:} $\forall c \in \mathcal{C}\:, \ \quad \mathrm{old\_label}(c) \in \mathcal(L)$
\ForAll{ $c \in \mathcal{C}$}
	\State $\mathit{Neighb} \gets \mathit{k\_nearest\_neighbour}(k, c, \mathit{max}\_\mbox{dist.})$
	\State $n = \mathit{card(Neighb)}$ 
	\State $ \mathit{new\_label}[c] \gets  \mathit{old\_label}[c]$
	\ForAll{ $ \mathit{class} \in  \mathcal(L)$}
		\State $ N_{c} \gets \{x|x \in {Neighb}, \mathit{old\_label}[x]=class\}$
		\If{ $\mathit{card}(N_c) > \frac{n}{2}$}
			\If{ $\sum_{x \in N_{c}} \mathit{area}(x) > \frac{1}{2}(c)}$, 
			\State $\mathit{new\_label}[c] \gets \mathit{class}$
			\EndIf
			
			     break
		\EndIf
	\EndFor
\EndFor
\end{algorithmic}
\end{algorithm}

\smallskip
\noindent {A3. \textit{Grouping by confidence voting.}} \\
The classifier confidence helps to maintain the decision. Based on the idea of grouping via nearest neighbours in addition with some specific constraints, we examine the confidence of the nearest neighbour of a selected pseudo-word. If the latter is stronger than that of the pseudo-word, then it takes the neighbourhood class. A Gaussian or polynomial law can weight the neighbour confidence by its distance to the pseudo-word.

\section{Experiments}\label{sec: exp}
\subsection{Dataset and evaluation metric}
\noindent\textit{\textbf{Dataset.}} To perform the tests, we have selected 75 documents for learning and a 300 documents for testing. As a reminder, these samples are taken from the real-world industrial problem.

\bigskip
\noindent\textit{\textbf{Evaluation metric.}} Our evaluation of $\cal H\& \cal P$ separation is performed according to the measure proposed by~\cite{Shafait}. All test documents have been perfectly labelled at pixel level, where performance is evaluated in terms of recognition rate.
\begin{eqnarray}
\mbox{Recognition rate} = \frac{\mbox{$\#$ of pixels correctly labelled}} {\mbox{$\#$ of pixels used}}.
\end{eqnarray} 

\subsection{Results and analysis} 
Table~\ref{tab:regroupe} shows recognition rates for four grouping methods. The $k$-NN uses $k=2$. The methods' confidence use respectively $f_{\mathit{gauss}}$,
$f_{\mathit{poly2}}$ and $f_{\mathit{poly4}}$ as weighting functions.

\begin{eqnarray}
f_{\mathit{gauss}}(\mathit{conf},\mathit{dist}) = conf \times \exp{\left(-\dfrac{10^{-3}*\mathit{dist}^2}{\mathit{conf}^2} \right)} \\
f_{\mathit{poly2}}(\mathit{conf},\mathit{dist}) = -5\cdot 10^{-4} {\left(\dfrac{  \mathit{dist}-1}{\mathit{conf}} \right) }^2 + \mathit{conf} \\
f_{\mathit{poly4}}(\mathit{conf},\mathit{dist}) = -10^{-6} {\left(\dfrac{  \mathit{dist}-1}{\mathit{conf}} \right) }^4 + \mathit{conf}
\end{eqnarray}
Based on reported results in Table~\ref{tab:regroupe}, we observe the following:
\begin{enumerate}\itemsep-0.5em
\item We note that the classification by $k$-NN provides better results as expected the recognition rate of double smearing i.e., segmentation without re-grouping. In contrast, methods based on confidence degrades performance. This is mainly due to the fact only local vicinity (a single neighbour) is taken into account, that makes misclassification possible. 
\item In our study, we have found that handwritten mixes with printed and other cases where grouping changes the isolated handwritten annotations label (e.g., a figure or a symbol). In this situation, we are required more contextual information including the better interpretation, which is beyond the scope of current work. 
\end{enumerate}
\begin{table}[hbp]
\centering
\small
\renewcommand{\tabcolsep}{0.5em}
\caption{Evaluation of four grouping methods.}\label{tab:regroupe}
\begin{tabular}{lcccc} 
  \hlinewd{1.5pt} Recognition rate & Hand. & Print. & Noise & Average \\ 
  \hlinewd{1.5pt} Double smearing            & 96.1 & 98.5 & \textbf{35.7} & 89.48 \\
  $k$-NN                    & 93.4 & 98.3 & 27.3 & 89.54 \\
  $k${NN} with constraints  & \textbf{99.3} & \textbf{99.0} & 27.9 & \textbf{90.68} \\
  Gaussian confidence    & 94.5 & 97.7 & 27.2 & 87.49 \\
  Poly confidence2 \& 4    & 93.5 & 97.7 & 14.2 & 86.06 \\
  \hlinewd{1.5pt}
  \end{tabular}
\end{table}
%%%%%%%%%%%%%%%%%%%%%%%%%%%%%%%%%%%%%%%%%%%%%%%%%%%%%%%
\begin{figure*}[htbp]
    \centering
        \fbox{\includegraphics[height = 5.5cm, width = 4cm]{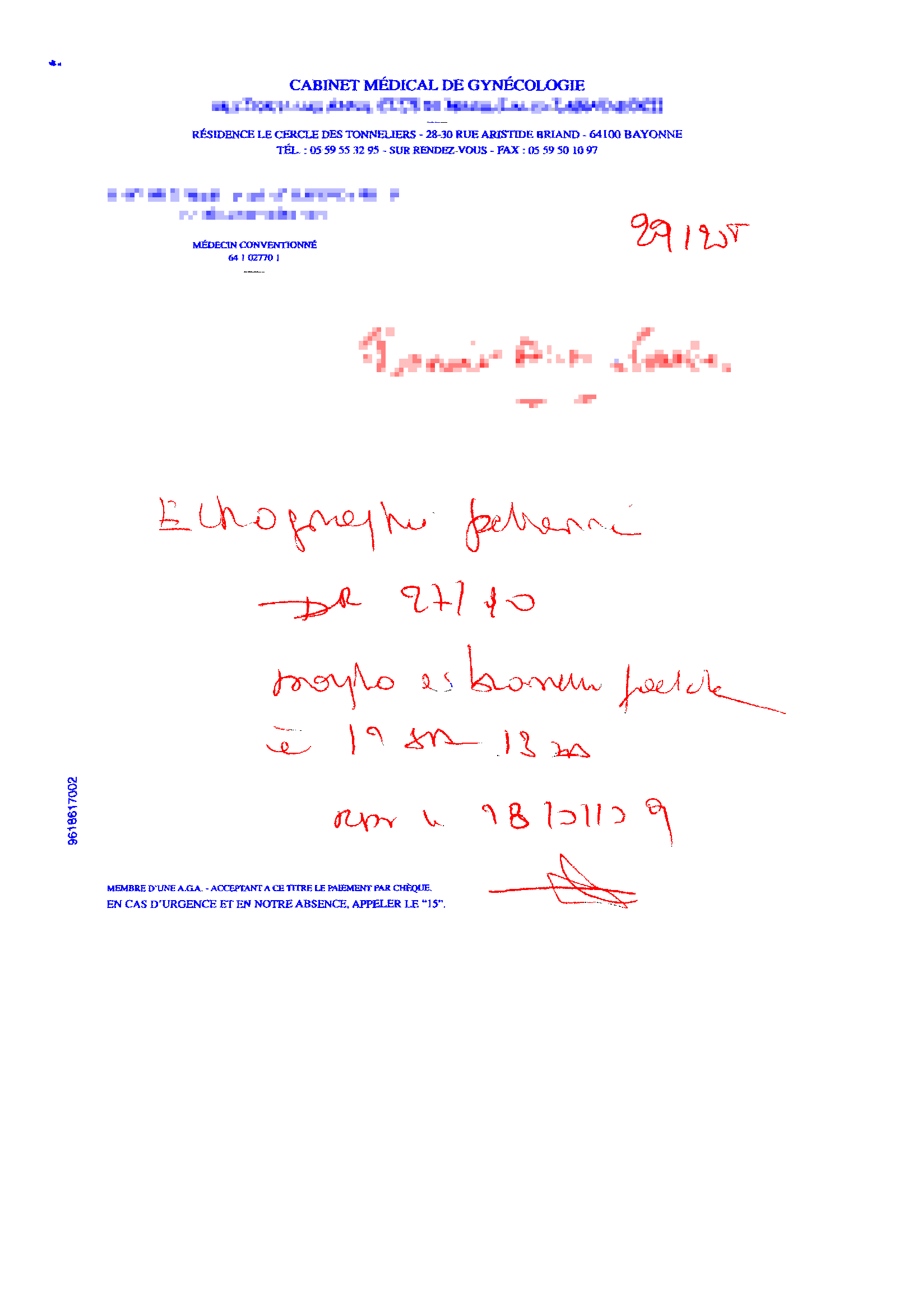}}
        \fbox{\includegraphics[height = 5.5cm, width = 4cm]{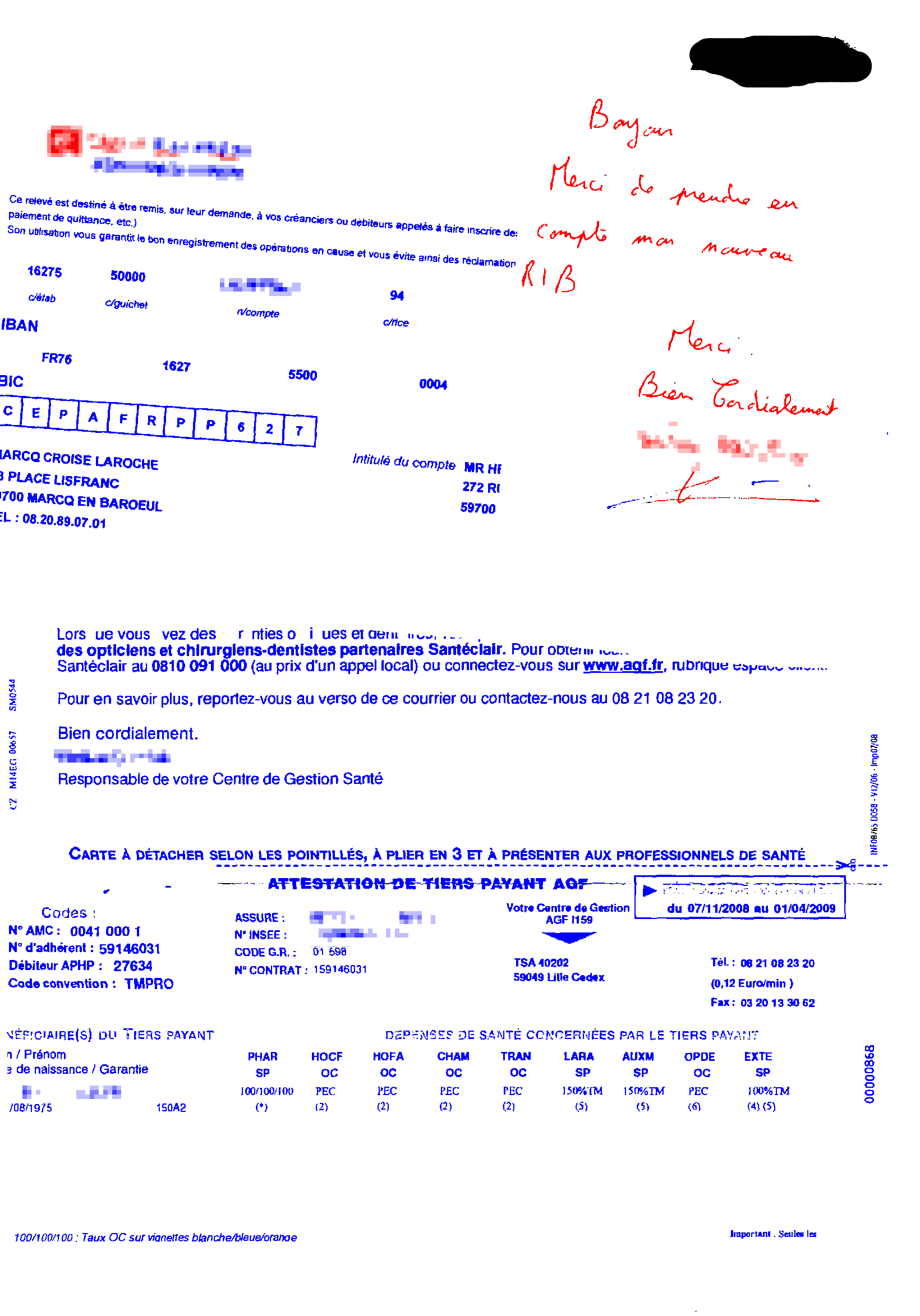}}
        \fbox{\includegraphics[height = 5.5cm, width = 4cm]{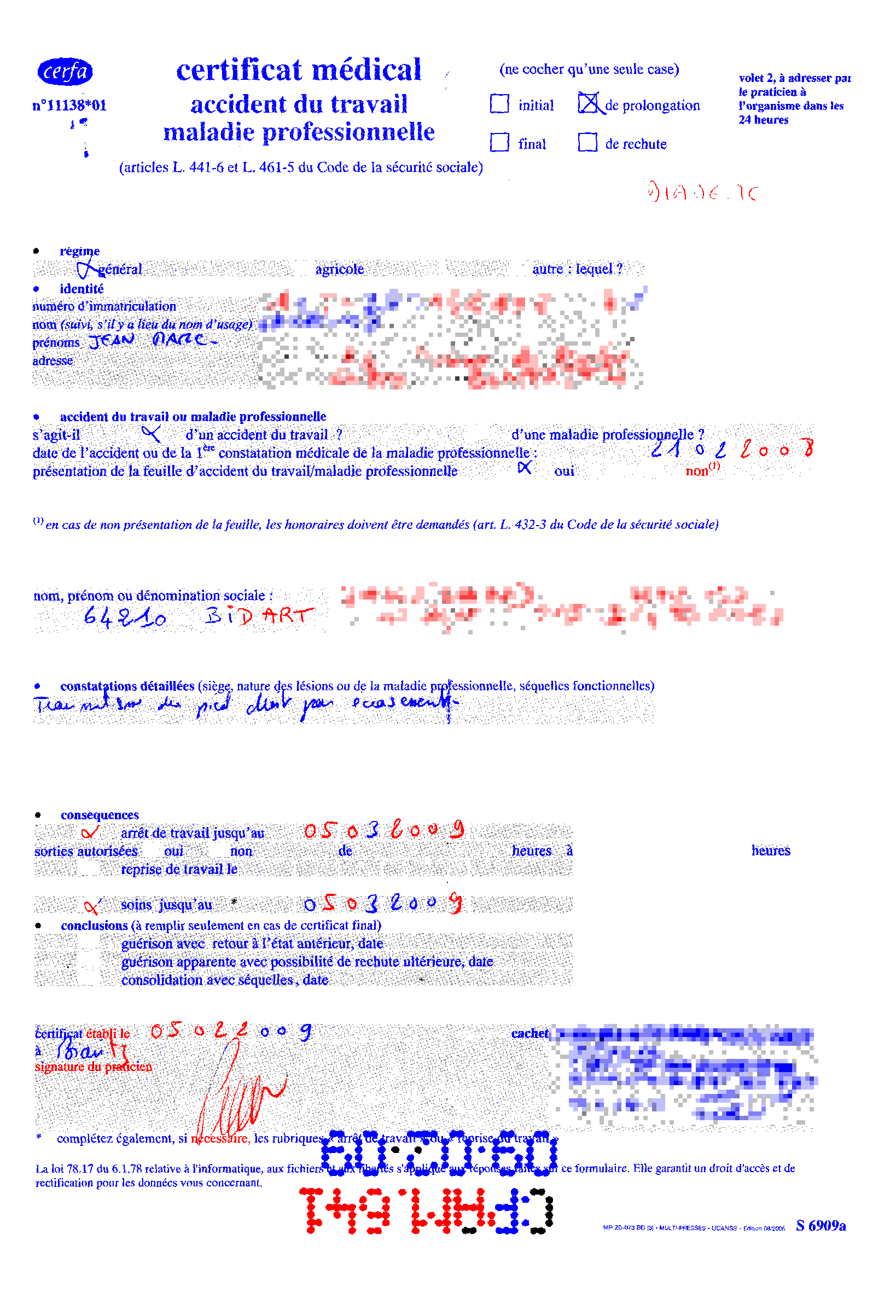}}
        \fbox{\includegraphics[height = 5.5cm, width = 4cm]{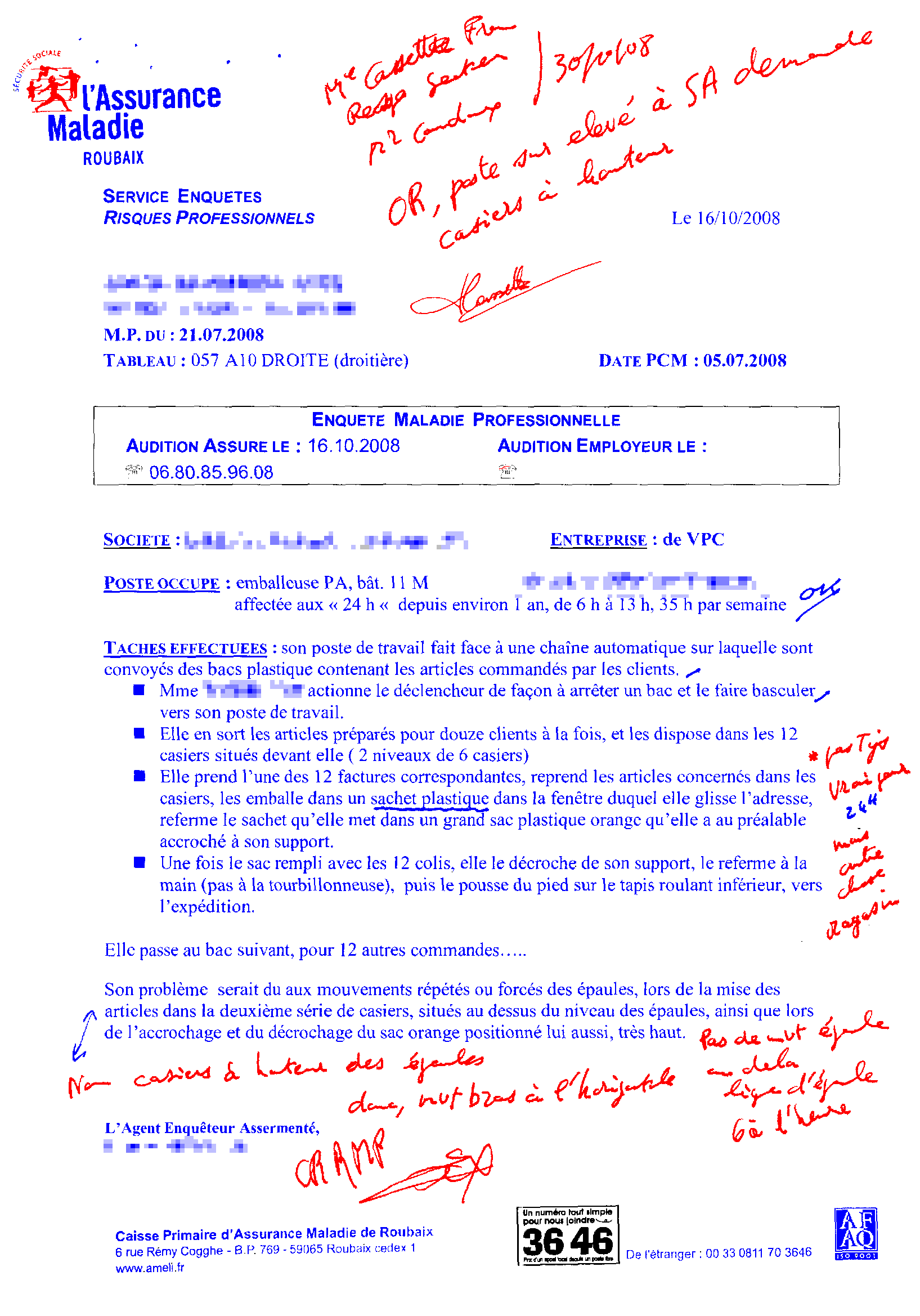}}
        \caption{A few examples of $\cal H\& \cal P$ text separation, illustrating the robustness of the proposed approach.}\label{exampl}
\end{figure*}
 
\begin{figure}[tbp]
    \centering
        \includegraphics[width=0.8\linewidth]{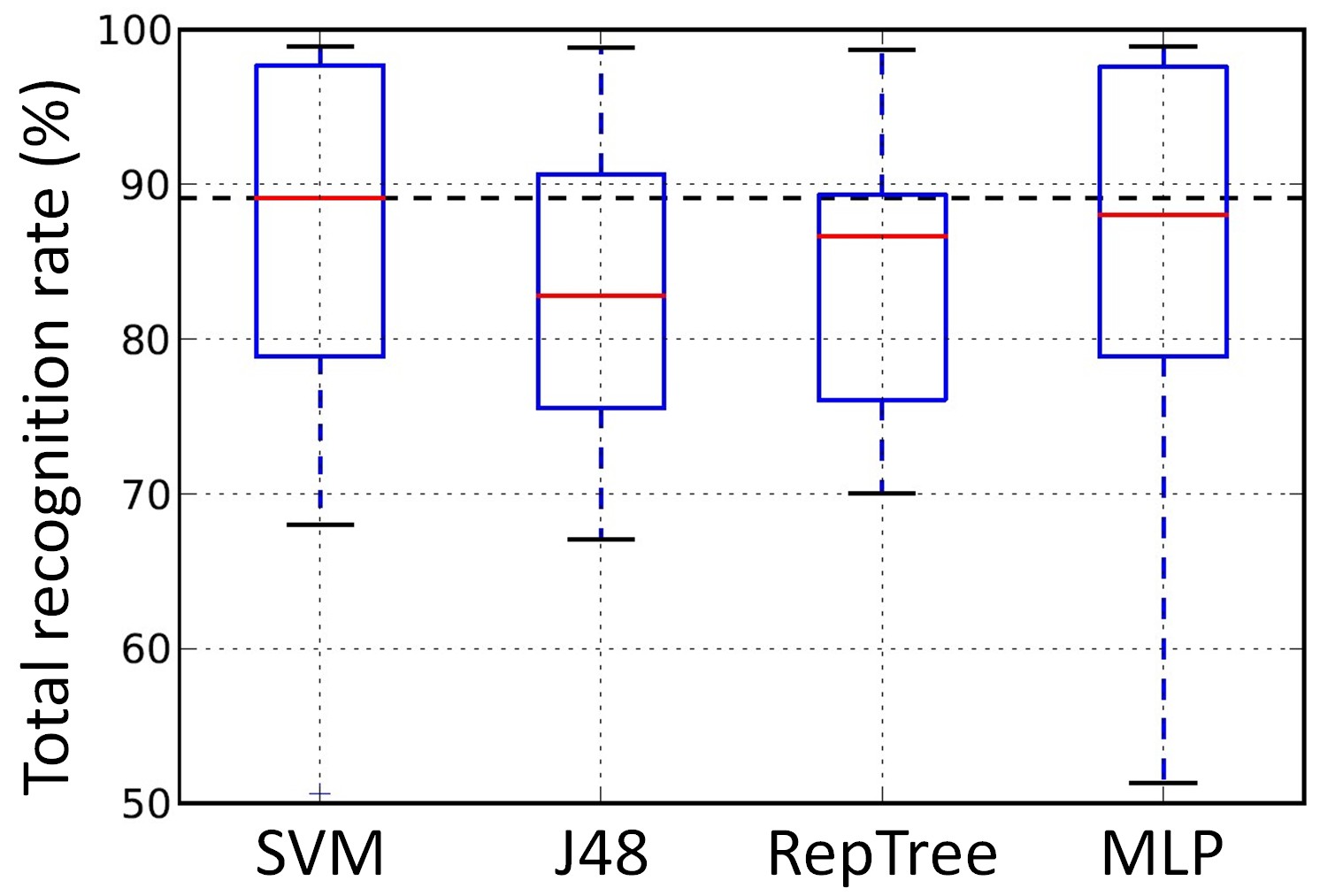}
    \caption{Evaluation of four classifiers}
    \label{fig:compclassif}
\end{figure}

On the whole, for visual understanding, we provide a few examples of $\cal H\& \cal P$ text separation in Fig.~\ref{exampl}. Furthermore, Fig.~\ref{fig:compclassif} shows a comparison between four classifiers: SVM, Tree C4.5  (J48 implementation), REPTree and NN. In this comparison, we have found that SVM performs the best, by providing marginal difference with NN. This means that MLP can still be applied. 
%%%%%%%%%%%%%%%%%%%%%%%%%%%%%%%%%%%%%%%%%%%%%%%%%%%%%%
%%%%%%%%%%%%%%%%%%%%%%%%%%%%%%%%%%%%%%%%%%%%%%%%%%%%%%%%%%
 
\balance
\section{Conclusion}\label{sec: concl}
In this paper, we have presented an approach to separate handwritten and machine-printed text from a scanned document in addition to the noise. The method is based on a double smearing technique to obtain the pseudo-words. These serve as a basis for classification. For these words, descriptors are extracted where they all have a linear complexity with the number of pixels. Descriptors are then fed into a multi-class SVM with a Gaussian kernel which provides the first label of each pseudo-word. A second analysis is carried out by studying the local vicinity of each pseudo-word that can change label if the neighbours are from another class. This integration allows context to correct several possible errors. In our test, we have found that the method is $k$-NN with constraints where kd-tree has been used.

Considering our small learning database, the results are fairly encouraging. This will certainly forecast an appropriate commercial application. Based on our reported results, a long-term approach about incremental learning is one of the further issues.

\section*{Acknowledgements}
The authors would like to thank Didier Grzejszczak and Yves Rangoni for their help and Herv\'e Locteau for his high level implementation to make the concept commercially useful. The work has been done during their stay with us at LORIA - Universit\'e de Lorraine. %Besides, the authors would also like to thank their collaborator - ITESOFT, Parc d’Andron, Le S\'equoia, 30470, Aimargues, France.   


\begin{thebibliography}{99}

\bibitem{Kang} 
Kang W.-X., Yang Q.-Q., Liang R.-P., The Comparative Research on Image Segmentation Algorithms, in: \textit{Proceedings of the ECTS}, 2009, pp. 703-707.
  
\bibitem{Chanda10}
S. Chanda, K. Franke, and U. Pal, Structural handwritten and machine print classification for sparse content and arbitrary oriented document fragments, in: \textit{Proceedings of SAC}, 2010, pp. 18-22.

\bibitem{setlur11}
Peng, X., Setlur, S., Govindaraju, V., and Sitaram, R., Handwritten text separation from annotated machine printed documents using markov random fields. \textit{IJDAR}, 16(1): 1-16, 2011.

\bibitem{Kandan} 
Kandan R., N.Kumar R., Arvind K.~R., Ramakrishnan A.~G., A robust two level classification algorithm for text localization in documents, in: \textit{Proceedings of the Advances in visual computing}, 2007, pp. 96-105.

\bibitem{Zheng} 
Zheng Y., Li H., Doermann D., The segmentation and identification of handwriting in noisy document images, in: \textit{Proceedings of DAS}, 2002, pp. 95-105.





\bibitem{Chinnasarn}
Chinnasarn K., Rangsanseri Y., Thitimajshima P., Removing Salt-and-Pepper Noise in Text/Graphics Images, {\em The Asia-Pacific Conference on Circuits and Systems}, 1998, pp. 459-462.

\bibitem{vanBeusekom} 
van Beusekom J., Shafait F., Breuel T.~M., Combined orientation and skew detection using geometric text-line modeling, in: \textit{Proceedings of the ICDAR}, 2010, pp. 79-92.

\bibitem{daSilva}
da~Silva L.~F., Conci A., Sanchez A., Automatic Discrimination between Printed and Handwritten Text in Documents, in: \textit{Proceedings of the Brazilian Symposium on
  CGIP}, 2009, pp. 261-267.

\bibitem{Vapnik}
V. N. Vapnik. The nature of statistical learning theory. Springer-Verlag New York, Inc., New York, NY, USA, 1995.

\bibitem{Weston} 
Weston J., Watkins C., Multi-class Support Vector Machines, {\em Technical report, Royal Holloway, University of London}, 1998.

\bibitem{Mayoraz} 
Mayoraz E., Alpaydin E., Support Vector Machines for Multi-class
  Classification, in: \textit{Proceedings of the ANN},  1999, pp. 833-842.

\bibitem{Shafait} 
Shafait F., Keysers D., Breuel T.~M., Performance Evaluation and
Benchmarking of Six-Page Segmentation Algorithms, \textit{IEEE-PAMI}, 30(6):941-954, 2008.

\end{thebibliography}
\end{document}